%% file: main.tex
\documentclass[10pt,twocolumn,letterpaper]{article}

\usepackage[pagenumbers]{cvpr} % To force page numbers, e.g. for an arXiv version


\usepackage{algorithm}
\usepackage[noend]{algpseudocode}
\usepackage{graphicx}
\usepackage{float}
\usepackage{overpic}
\usepackage{wrapfig}
\usepackage{caption}
\usepackage[T1]{fontenc}

\definecolor{cvprblue}{rgb}{0.21,0.49,0.74}
\usepackage[pagebackref,breaklinks,colorlinks,allcolors=cvprblue]{hyperref}

\def\confName{CVPR}
\def\confYear{2025}

\title{MoDiT: Learning Highly Consistent 3D Motion Coefficients with Diffusion Transformer for Talking Head Generation}

\author{
Yucheng Wang and Dan Xu\\
The Hong Kong University of Science and Technology\\
{\tt\small ywangls@connect.ust.hk, danxu@cse.ust.hk}
}

\begin{document}
% \maketitle
\twocolumn[{%
\renewcommand\twocolumn[1][]{#1}%
\maketitle
\begin{center}
    \centering
    \captionsetup{type=figure}
    \vspace{-18pt}    \includegraphics[width=0.99\textwidth]{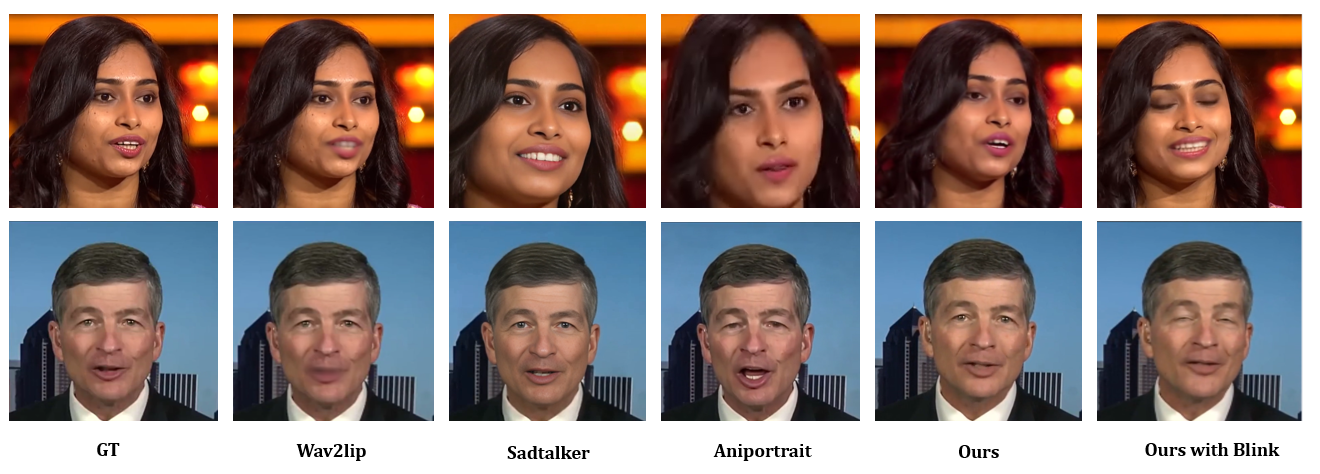}
    \vspace{-5pt}
    \captionof{figure}{MoDiT learns highly consistent 3D Motion Coefficients~\citep{3dmm} (3DMM) with diffusion transformer and predicts accurate flow based on the previously learnt 3DMM, which enhances the consistency of the talking head video. We aim to address temporal jittering caused by weak temporal constraints leading to frame inconsistencies, identity drift resulting from insufficient 3D information extraction compromising facial identity preservation, and unnatural blinking behavior due to inadequate modeling of realistic blink dynamics. 
    }
    \label{fig:teaser}
    % \vspace{-10pt}
\end{center}%
}]

\begin{abstract}
\input{Content/Chapter/Abstract}
\vspace{-15pt}
\end{abstract}

\section{Introduction}
\input{Content/Chapter/Introduction}

\section{Related Work}
\input{Content/Chapter/RelatedWork}

\section{The Proposed MoDiT Method}
\input{Content/Chapter/Methods}

\section{Experiment}
\input{Content/Chapter/Experiments}

\section{Conclusion}
\input{Content/Chapter/Conclusion}
\newpage
\small
\bibliographystyle{ieeenat_fullname}
\bibliography{main}

% \section{Appendix}
% \input{Content/Chapter/Appendix}
\end{document}

%% file: Content/Chapter/Abstract.tex
Audio-driven talking head generation is critical for applications such as virtual assistants, video games, and films, where natural lip movements are essential. Despite progress in this field, challenges remain in producing both consistent and realistic facial animations. Existing methods, often based on GANs or UNet-based diffusion models, face three major limitations: (i) temporal jittering caused by weak temporal constraints, resulting in frame inconsistencies; (ii) identity drift due to insufficient 3D information extraction, leading to poor preservation of facial identity; and (iii) unnatural blinking behavior due to inadequate modeling of realistic blink dynamics. To address these issues, we propose MoDiT, a novel framework that combines the 3D Morphable Model (3DMM) with a Diffusion-based Transformer. Our contributions include:
(i) A hierarchical denoising strategy with revised temporal attention and biased self/cross-attention mechanisms, enabling the model to refine lip synchronization and progressively enhance full-face coherence, effectively mitigating temporal jittering.
(ii) The integration of 3DMM coefficients to provide explicit spatial constraints, ensuring accurate 3D-informed optical flow prediction and improved lip synchronization using Wav2Lip results, thereby preserving identity consistency.
(iii) A refined blinking strategy to model natural eye movements, with smoother and more realistic blinking behaviors.
Please visit our repository.

%% file: Content/Chapter/Introduction.tex
\par Audio-driven talking head generation \citep{chung2017lip,zhou2021pose,hong2025audio} plays a crucial role in applications such as virtual assistants, video games, and movie productions. The goal of this task is to generate lifelike lip movements that are precisely synchronized with the driving audio. Achieving this synchronization allows digital humans to effectively convey the tone and rhythm of speech, creating a more engaging interaction experience for users.

\par Despite promising advancements in audio-driven talking head generation~\citep{xu2024hallohierarchicalaudiodrivenvisual, sadtalker,tang2025human,ma2023dreamtalk}, several critical challenges persist in producing highly consistent results. (i) The recent methods~\citep{rombach2022high,shen2023difftalk,du2023dae} often face problems like jittering and identity inconsistency. We argue that UNet-based diffusion models struggle to maintain consistency from both spatial and temporal perspectives. Spatially, the pre-trained Variational Autoencoder (VAE) encoder~\citep{vae} fails to extract explicit 3D information, leading to insufficient spatial reference, for example, Aniportrait~\citep{wei2024aniportrait} in Figure~\ref{fig:teaser}. Temporally, despite recent advances with temporal modules~\citep{xu2024hallohierarchicalaudiodrivenvisual} aimed at improving consistency, there are still insufficient constraints on how latent sequences are expressed over time, which affects coherence. (ii) Although diffusion models present a potential solution by gradually removing noise to capture the entire distribution, current techniques still rely on GRUs or CNNs for sequential processing. These structures struggle with extensive context compared to Transformers, which results in temporal inconsistency. (iii) The correlation between lip movements and the driving audio is stronger than that between facial motions (like head movements and blinkings) and driving audio. Thus, it is an ill-posed problem to find mappings between lip movements, facial motions, and driving audio. In this situation, traditional methods with deterministic regression often fail to capture the joint distribution due to learning a single mapping directly. While Generative Adversarial Networks (GANs)~\citep{goodfellow2020generative} have shown promise, they still encounter the problem of accommodating a wide range of facial expressions and instability in the training stage due to learning a mapping directly from a noise distribution to the data distribution. An example is Wav2Lip~\citep{wav2lip} shown in Figure~\ref{fig:teaser}.

\par  Our work - MoDiT (shown in Figure~\ref{fig:figure1}) presents a novel approach to integrating the 3D Morphable Model (3DMM)~\citep{3dmm} with a Diffusion-based Transformer framework. The 3DMM offers explicit constraints during frame rendering, which can reduce spatial inconsistency. During the diffusion denoising process, we utilize latent variables extracted from the audio signal along with the initial 3DMM to establish temporal and spatial conditions. (i) By developing a specialized strategy that incorporates biased conditional self/cross-attention, we allow the diffusion model to focus on different levels during various denoising stages, starting from the lip region and expanding to the entire face. This hierarchical approach helps balance lip synchronization and the naturalness of overall expressions while reducing computational complexity. (ii) We use accurate 3DMM coefficients to predict accurate flow and use the results of Wav2Lip~\citep{wav2lip} to ensure better Lip-Sync in the rendering pipeline. (iii) In addition to this, we provide a new strategy to make blinking smoother.  

\par We used data from VFHQ~\citep{wang2022vfhqhighqualitydatasetbenchmark} and HDTF~\citep{hdtf} for training, and tested same-identity and cross-identity on both datasets, as shown in Table~\ref{tab:aud_driven}. To demonstrate the scalability of our model, we evaluated it on a completely out-of-distribution dataset, as shown in Table~\ref{tab:tableHQ}.  Considering the Hallo~\citep{xu2024hallohierarchicalaudiodrivenvisual} is trained on a larger private dataset, we argue that our method achieves SOTA-comparable results in temporal and spatial consistency.   

\par In summary, our work presents a novel approach to integrating the 3D Morphable Model (3DMM) with a Diffusion-based Transformer framework for audio-driven talking head generation. Our contributions include: (i) A hierarchical denoising strategy with revised temporal attention and biased self/cross-attention mechanisms, enabling the model to refine lip synchronization and progressively enhance full-face coherence, effectively mitigating temporal jittering.
(ii) The integration of 3DMM coefficients to provide explicit spatial constraints, ensuring accurate 3D-informed optical flow prediction and improved lip synchronization using Wav2Lip results, thereby preserving identity consistency.
(iii) A refined blinking strategy to model natural eye movements, with smoother and more realistic blinking behaviors.

%% file: Content/Chapter/RelatedWork.tex
\par
\textbf{2D-based Talking Head Generation} 
Early methods for talking head generation consider facial motion extracted from a given face video sequence as driving signals to generate a talking head video~\cite{fomm,hong2022depth,hong2023dagan++,hong2023implicit,zhao2022thin,zhao2025synergizing}. 
For more convenient interaction, the audio driving methods use encoder-decoder models to map audio to visual features, enabling realistic lip movements synchronized with speech~\citep{chung2017you, zhou2019talking,song2018talking}. Techniques like disentangling appearance and semantic features improved lip-sync accuracy, while evaluation networks ensured synchronization \citep{song2018talking, zhou2019talking, wav2lip}. Generative models (GANs, VAEs) refined these approaches by focusing on the mouth region or editing full frames to include head and shoulder movements \citep{chen2019hierarchical, thies2020neural, sun2022masked}. However, the lack of explicit 3D structural data often caused inconsistencies like unnatural poses or misaligned features \citep{suwajanakorn2017synthesizing, liu2023moda, ji2021audio}.

\noindent \textbf{3D-based Talking Head Generation}
3D-based methods advanced facial animation by capturing head motions, emotional expressions, and facial details for lifelike results~\citep{dvp, lsp, evp}. Models like AudioDVP~\cite{audiodvp}, NVP~\cite{thies2020neural}, and AD-NeRF~\cite{adnerf} reenact facial expressions dynamically in 3D space, ensuring consistency across head poses and lighting. By integrating 3D geometry~\cite{ricci2018monocular} with neural rendering, these methods achieve superior realism and stability compared to 2D approaches, which struggle with extreme poses and spatial inconsistencies~\citep{sadtalker}.

\begin{figure*}[!t]
    \centering    \includegraphics[width=1.0\linewidth]{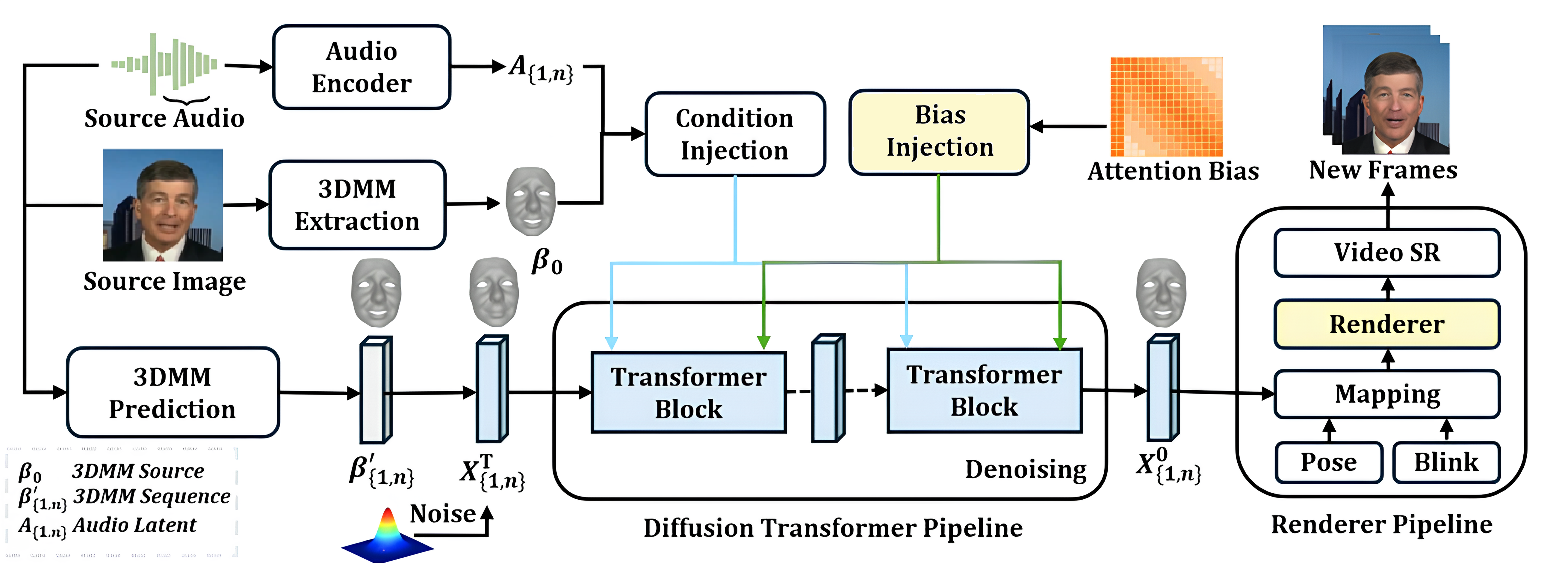}
    \vspace{-10pt}   \caption{Overview of the Diffusion Transformer Pipeline, showcasing the denoising stages with temporal and spatial condition injection. It uses 3DMM for spatial conditions and an audio encoder for temporal conditions. The diffusion transformer denoises 3DMM sequences, which are then processed through pose prediction, mapping, and rendering to new frames.}   \label{fig:figure1}
    \vspace{-10pt}
\end{figure*}

\noindent \textbf{Talking Head Generation with Diffusion Methods} 
Diffusion models have revolutionized talking head synthesis by iteratively refining images for high-quality animations. Early models like Diffused Heads~\citep{stypulkowski2024diffused} and GAIA~\citep{ he2023gaia} set the foundation, while DreamTalk~\citep{ma2023dreamtalk} and VividTalk~\citep{sun2023vividtalk} introduced more expressive facial animations. Recent advancements, including VASA-1~\citep{xu2024vasa}, EMO~\citep{tian2024emo}, AniTalker~\citep{liu2024anitalker}, and Hallo~\citep{xu2024hallohierarchicalaudiodrivenvisual}, further enhance emotional expressiveness, long video handling, and character customization, pushing the boundaries of realistic synthesis.~ACTalker~\cite{hong2025audio} proposes a mamba-based condition Diffusion framework for multi-modal talking head generation. While these methods achieve impressive results in expression modeling, temporal coherence, and character customization, they often lack explicit spatial constraints for lip synchronization and fail to address natural eye movements comprehensively.

%% file: Content/Chapter/Methods.tex
\subsection{Preliminary of 3D Face Model}
We consider the space of the predicted 3D Morphable Models~(3DMMs) as our intermediate representation.
In 3DMM, the 3D face shape $\mathbf{S}$ can be decoupled as:
\begin{equation}
\mathbf{S} = \overline{\mathbf{S}} + \mathbf{\alpha} \mathbf{U}_{id} + \mathbf{\beta} \mathbf{U}_{exp},
\end{equation}
where \(\overline{\mathbf{S}}\) represents the average 3D face shape, and \(\mathbf{U}_{id}\) and \(\mathbf{U}_{exp}\) are the orthonormal bases for identity and expression in the LSFM morphable model \citep{3dmm}. The coefficients \(\mathbf{\alpha} \in \mathbb{R}^{80}\) and \(\mathbf{\beta} \in \mathbb{R}^{64}\) capture identity and expression, respectively. To enhance lip synchronization during generation, we model only the expression parameters  \(\mathbf{\beta} \in \mathbb{R}^{64}\).

\subsection{Diffusion Models for Talking Head Generation}
In the denoising phase~\citep{song2022denoisingdiffusionimplicitmodels}, our model is trained to reverse the diffusion process, converting random noise back into a real data distribution during inference. This denoising process can be expressed as:
\begin{eqnarray}
& p_{\theta}\left(\mathbf{x}_{t-1} \mid \mathbf{x}_{t}\right) = \mathcal{N}\left(\mathbf{x}_{t-1} ; \mu_{\theta}\left(\mathbf{x}_{t}, t\right), \Sigma_{\theta}\left(\mathbf{x}_{t}, t\right)\right) \nonumber\\ &= \mathcal{N}\left(\mathbf{x}_{t-1} ; \frac{1}{\sqrt{\alpha_{t}}}\left(\mathbf{x}_{t}-\frac{\beta_{t}}{\sqrt{1-\bar{\alpha}_{t}}} \mathbf{\epsilon} \right),
\frac{1-\bar{\alpha}_{t-1}}{1-\bar{\alpha}_{t}} \beta_{t}\right),
\end{eqnarray}
where $\mathbf{\epsilon} \sim \mathcal{N}(\mathbf{0},\mathbf{I})$, $\alpha_{t}=1-\beta_{t}$, $\bar{\alpha}_{t}=\prod_{i=1}^{t} \alpha_{i}$ and $\theta$ specifically denotes parameters of a neural network learning to denoise. The objective of training is to maximize the likelihood of the observed data \( p_{\theta}(\mathbf{x}_{0}) = \int p_{\theta}(\mathbf{x}_{0:T}) \, d\mathbf{x}_{1:T} \) by optimizing the evidence lower bound (ELBO). During training, the denoising network \(\mathbf{\epsilon}_\theta(\cdot)\) aims to reconstruct \(\mathbf{x}_0\) from any noised input \(\mathbf{x}_t\) by predicting the added noise \(\epsilon \sim \mathcal{N}(\mathbf{0}, \mathbf{I})\) and minimizing the noise prediction error:

% \end{align}
\begin{eqnarray}
\mathcal{L}_{t} & = \mathbb{E}_{\mathbf{x}_{0}, \epsilon }\left[\left\|\mathbf{\epsilon}-\mathbf{\epsilon}_{\theta}\left(\sqrt{\bar{\alpha}_{t}}\mathbf{x}_0+\sqrt{1-\bar{\alpha}_{t}} \mathbf{\epsilon}, t\right)\right\|^{2}\right]. \label{equ:noise_loss}
\end{eqnarray}
To condition the model on additional context information \(\mathbf{c}\), such as audio, we incorporate \(\mathbf{c}\) into \(\mathbf{\epsilon}_\theta(\cdot)\) by replacing \(\mu_\theta\left(\mathbf{x}_t, t\right)\) and \(\Sigma_\theta\left(\mathbf{x}_t, t\right)\) with \(\mu_\theta\left(\mathbf{x}_t, t, \mathbf{c}\right)\) and \(\Sigma_\theta\left(\mathbf{x}_t, t, \mathbf{c}\right)\).

\subsection{Transformer Block with Biased Attention} 
\par Utilizing the conventional Transformer architecture~\citep{khan2021transformers}, the outputs from the self-attention layer ($\mathbf{A}_s$) and the cross-attention layer ($\mathbf{A}_c$) are determined as follows:
\begin{equation}
    \begin{aligned}
        \mathbf{A}_s &= \operatorname{Attention}\left(\mathbf{Q}_s, \left[ \mathbf{\beta}_0; \mathbf{K}_s\right], \left[\mathbf{\beta}_0;  \mathbf{V}_s\right]\right), \\
        \mathbf{A}_c &= \operatorname{Attention}\left(\mathbf{Q}_c, \left[\mathbf{A}_n;\mathbf{K}_c\right], \left[\mathbf{A}_n;\mathbf{V}_c\right]\right),
    \end{aligned}
    \label{equ:8}
\end{equation}
$\mathbf{Q}_s$, $\mathbf{K}_s$, and $\mathbf{V}_s \in \mathbb{R}^{T \times C}$ represent the query, key, and value features for self-attention, while $\mathbf{Q}_c$, $\mathbf{K}_c$, and $\mathbf{V}_c \in \mathbb{R}^{T \times C}$ are their counterparts for cross-attention. The square brackets indicate concatenation along the sequence dimension. 

\subsection{MoDiT Pipeline} 

As shown in Figure \ref{fig:figure1}, our pipeline is crafted to generate new video frames from a given source audio and a single source image. The audio and image are first pre-processed into an audio latent $\displaystyle A_{\{1,n\}}$ and an initial 3DMM $\mathbf{\beta}_0$, which are then input into a 3DMM prediction module. The Temporal and Spatial Conditions guide the subsequent Diffusion Transformer Pipeline, composed of Transformer Blocks in a denoising stage. These blocks handle expression predictions to create intermediate 3DMM states. Finally, a Mapping Net uses these outputs to predict poses, which are rendered into new frames by a Renderer.

\subsubsection{Transformer Block with Biased Attention} 
\begin{figure}[!t]
    \vspace{-15pt}
    \centering   \includegraphics[width=1\linewidth]{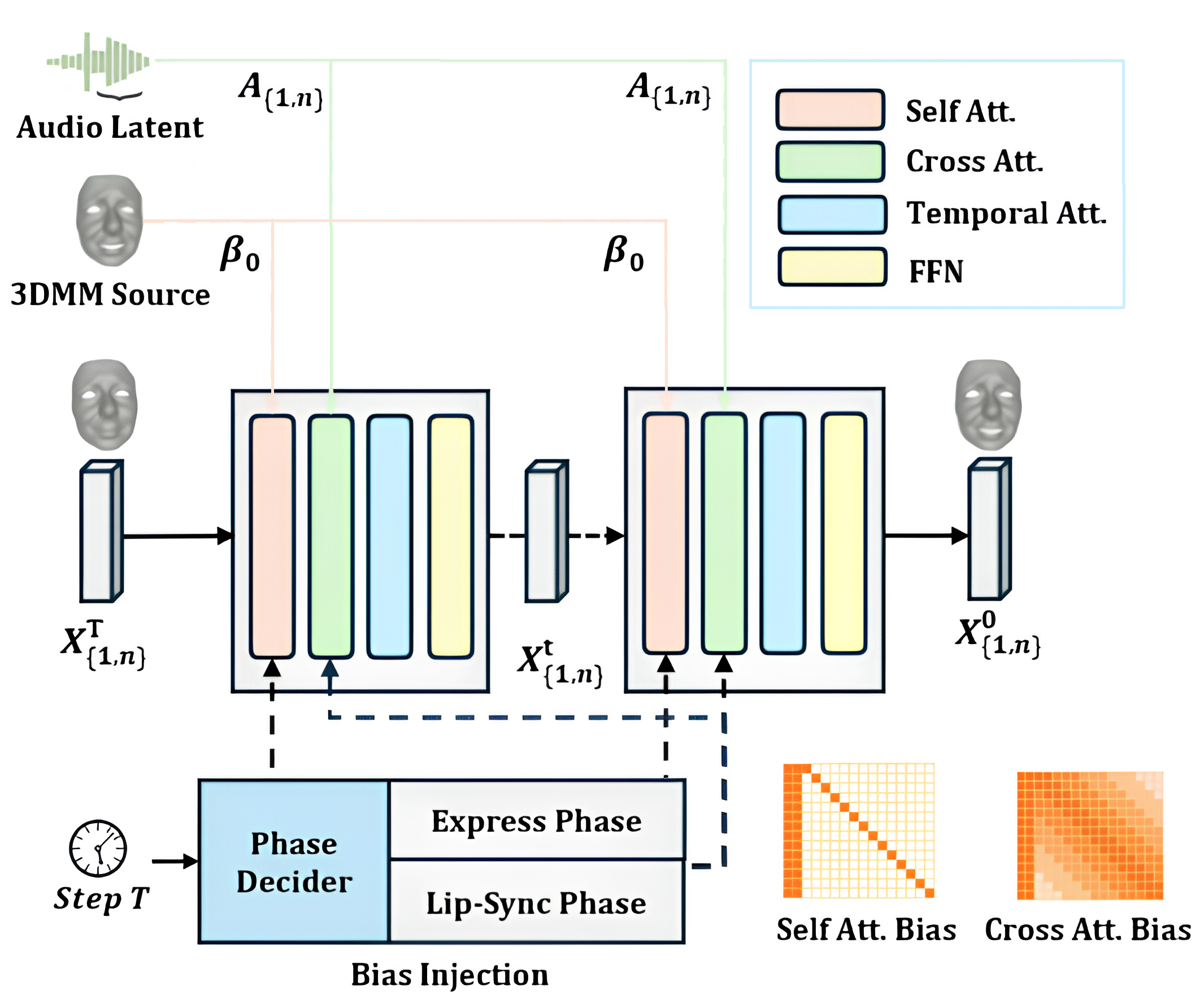}
    \vspace{-10pt}
    \caption{Illustration of structure details of the Transformer Block with the designed integrated Bias Injection Strategy, showing the use of spatial and temporal conditions to enhance attention mechanisms for improved feature extraction.}
    \vspace{-10pt}
    \label{fig:figure2}
\end{figure}

As shown in Figure \ref{fig:figure2}, the transformer block~\citep{khan2021transformers} in our method incorporates a multi-faceted attention mechanism to enhance temporal and spatial feature extraction. It begins with self-attention, which captures input noise latent $\displaystyle X_{\{1,n\}}^{t}$ with the condition of the initial 3DMM $\mathbf{\beta}_0$, followed by cross-attention to integrate audio condition $\displaystyle A_{\{1,n\}}$ as shown in Equation \ref{equ:8}. Temporal attention is then applied to refine the sequence over time. A feed-forward network (FFN) processes the attention outputs, producing predicted noise $\displaystyle \epsilon_{\{1,n\}}^{t}$. The block is modulated by a bias injection strategy (stated in 3.3.2), which introduces spatial and temporal biases through a phase decider that operates in the express and lip-sync phases, optimizing the model for tasks requiring precise temporal synchronization and spatial alignment. 

\begin{algorithm}[h]
\caption{Bias Injection Strategy} \label{alg:ddim_repaint}
\textbf{HyperParameter}: Latent Length $L$, Diffusion Steps $T$, Threshold ${t}_T$, Diagonal bias $\mathcal{M}_D$, Dispersed expression bias $\mathcal{M}_E$, Expression Attention $\mathcal{A}_E$. \\
\textbf{Output}: Generated expression $x_0$ 
\begin{algorithmic}[1]
    \State $x_t \sim \mathcal{N}(\mathbf{0}, \mathbf{I})$\Comment{$x_t \in R^{L \times C}$, where $C$ is channel size.}
    \For{$t=T, \dotsc, 1$}
        \vspace{1mm}
          \State $\epsilon \sim \mathcal{N}(\mathbf{0}, \mathbf{I})$ if $t > 1$, else $\epsilon = \mathbf{0}$
          \vspace{1mm}
          \State $ \hat{x}_0^t = 
          \frac{ x_t - \sqrt{1-\bar{\alpha}_t} \mathbf{\epsilon}_\theta(x_t, t)} {\sqrt{\bar{\alpha}_t}}
          $
          \vspace{1mm}
          \If{$t < {t}_T$}
    \State $\mathcal{A}_E = \mathcal{M}_D \odot \mathcal{A}_E + \mathcal{M}_E \odot \mathcal{A}_E $
\Else
    \State $\mathcal{A}_E = \mathcal{M}_E \odot \mathcal{A}_E $
\EndIf
          \vspace{1mm}
          \State $x_{t-1} =
          %%%%%%%%% DDIM %%%%%%%%%%
          \sqrt{\bar\alpha_{t-1}}  
          \hat{x}_0^t + 
          \sqrt{1-\bar{\alpha}_{t-1}} \mathbf{\epsilon}_\theta(x_t, t)
          $ 
          %%%%%%%%%%%%%%%%%%%    
          \State $x_t \sim \mathcal{N}(\sqrt{1-\beta_{t-1}} x_{t-1}, \beta_{t-1} \mathbf{I})$
          \vspace{1mm}
    \EndFor
    \State \textbf{return} {$x_0$}
\vspace{.04in}
\end{algorithmic}
\end{algorithm}

\subsubsection{Bias Injection Strategy} 

The Bias Injection Module, inspired by Diffspeaker~\citep{ma2024diffspeakerspeechdriven3dfacial}, is a novel element crafted to enhance the diffusion process by segmenting it into specific phases, each utilizing focused attention mechanisms. For details of the Bias Injection Strategy, refer to Algorithm~\ref{alg:ddim_repaint}. In the early stages of diffusion, when noise levels are high, the module prioritizes lip alignment using a diagonal bias. This bias helps concentrate computational resources on lip synchronization, ensuring precise alignment despite the presence of substantial noise. As the diffusion progresses and noise diminishes, the module shifts focus to the overall facial motions. During this later stage, a more dispersed expression bias is applied to facilitate smooth and natural transitions in facial expressions. This phased approach not only enhances computational efficiency but also significantly improves the lip synchronization rate, resulting in a harmonious blend of detailed lip movements and facial motions. For more details of the Bias Injection Strategy, refer to the Algorithm~\ref{alg:ddim_repaint}.

\subsubsection{Revised Temporal Attention for Jitter Removal} 
Temporal attention~\citep{xu2017jointlyattentivespatialtemporalpooling} is a fundamental mechanism for capturing dependencies across time steps by dynamically weighing past and future information. Traditionally, attention scores \( \alpha_{t} \) are computed using query and key vectors \( Q_{t} \) and \( K_{t} \) as:
\begin{equation}
\begin{aligned}
\alpha_{t} &= \text{softmax}\left(\frac{Q_{t}K_{t}^T}{\sqrt{d_k}}\right)
\end{aligned}
\end{equation}

\par While effective, this approach struggles to adapt to complex temporal variations, as it treats all time steps uniformly without incorporating additional contextual information. To address this limitation, we propose a novel enhancement by introducing a learned latent variable \( z_t \) that dynamically adjusts the attention scores based on the sequence context. Specifically, the initial attention scores \( s_t \) are refined through a transformation function implemented as an MLP. The enhanced attention mechanism becomes:
\begin{equation}
    \begin{aligned}
        s_t &= \frac{Q_t K_t^T}{\sqrt{d_k}},
        \tilde{s}_t &= s_t + f(z_t),
        \alpha_t &= \text{softmax}(\tilde{s}_t)
    \end{aligned}
\end{equation}

\par This modification allows the model to adaptively adjust attention weights based on temporal context, enabling it to better capture subtle patterns and variations over time.

\subsubsection{Blinking Module Design} 
\begin{figure}[!t]
    \centering   \includegraphics[width=0.8\linewidth]{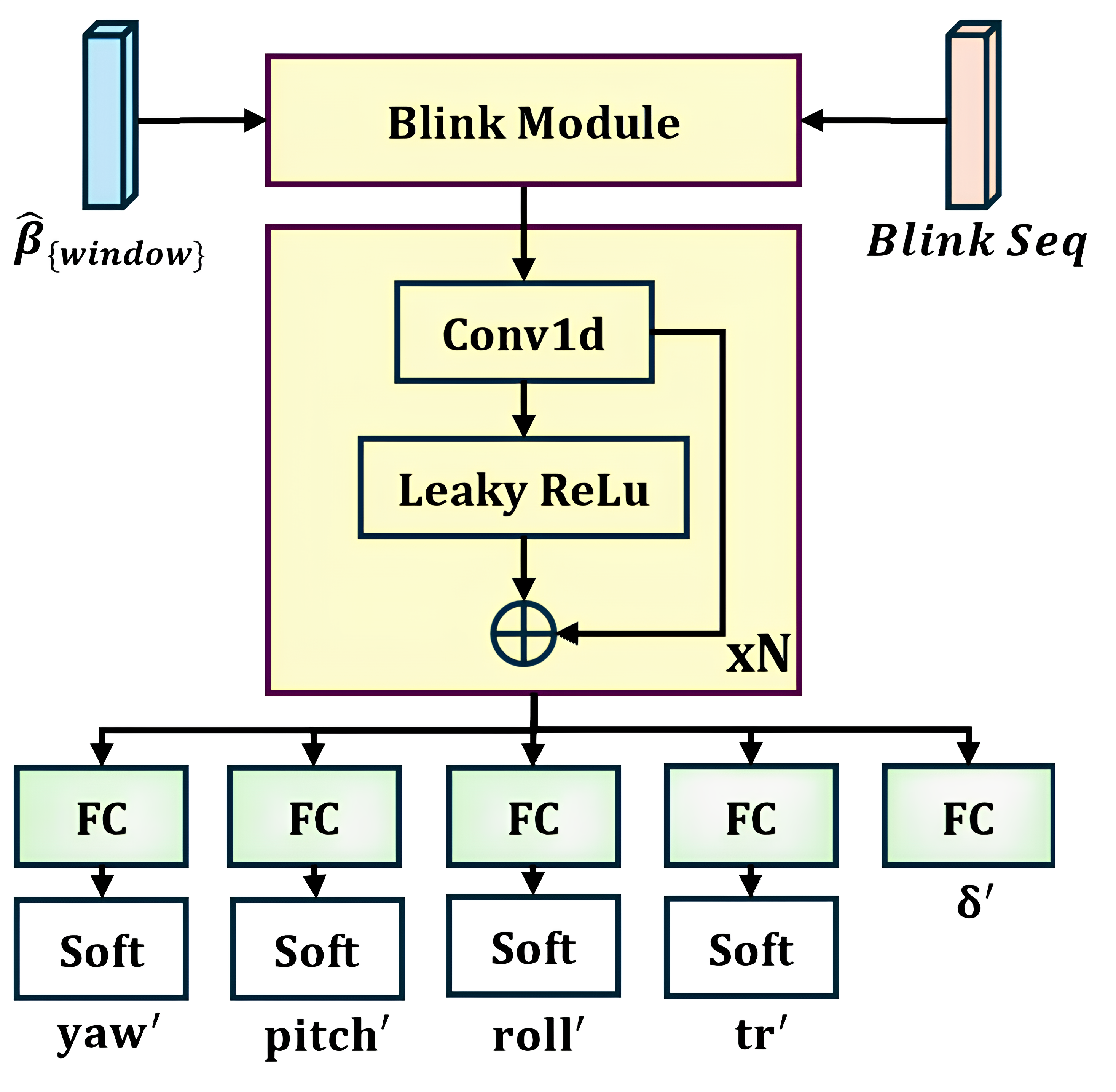}
    \vspace{-10pt}
    \caption{The mapping net structure for generating realistic eye blinks, featuring a Blink Module for head movement refinement.}
    \label{fig:mapping}
\end{figure}

The structure of the modified mapping net is shown in Figure \ref{fig:mapping}, which is essential for generating realistic eye blinks in video sequences. This module is designed to ensure that the 3DMM in the eye area effectively triggers blinks. The mapping net generates realistic eye blinks by employing a learnable blink sequence to determine the degree of eye closure for each frame, rather than relying on a simple binary signal.

The Blink Module fuses input features \(\hat{\beta}_{\{\text{window}\}}\) and blink sequence through multiple convolutional layers, each followed by Leaky ReLU activations, allowing the network to capture complex temporal dependencies and subtle nuances in blinking behavior. This architecture is iterated \(N\) times to refine the model's ability to generate natural blink sequences that can dynamically adjust blink intensity. Furthermore, the module is pre-trained using a cropped dataset focusing on eye movements, ensuring that the blink sequence accurately reflects the degree of eye closure. This pre-training step is critical as it can better align the blink sequence with the 3DMM parameters, enabling direct control over the eye area's 3D structure. Then, the network integrates fully connected (FC) layers followed by softmax activations, which refine outputs related to head movements such as yaw, pitch, roll, and translation (\(tr'\)). A parameter \(\delta'\) is included for further adjustments similar to SadTalker~\cite{sadtalker}.

\subsubsection{Renderer for Precise Flow Prediction} 
\begin{figure}[t!]
    \centering   \includegraphics[width=1\linewidth]{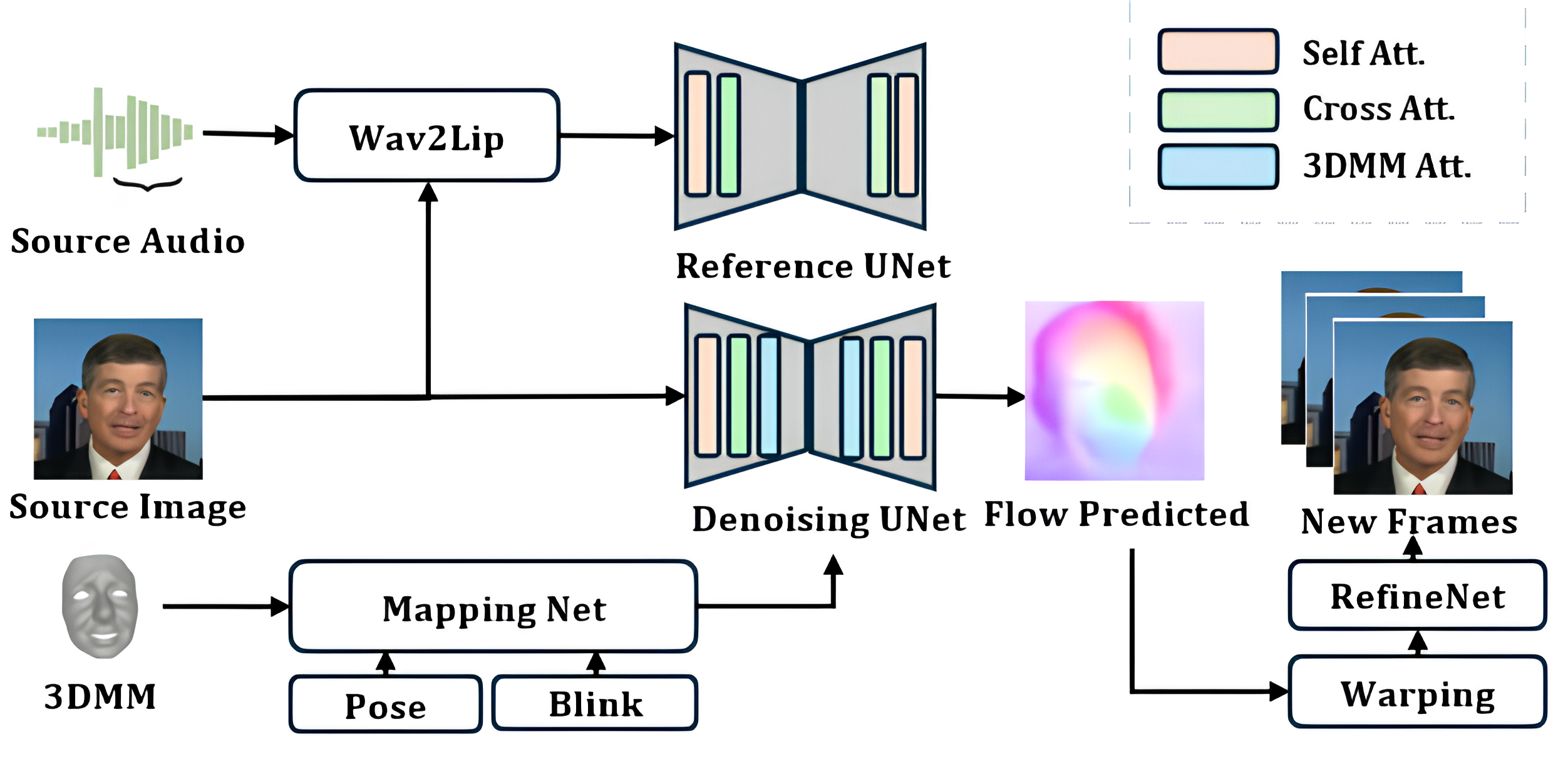}
    \vspace{-20pt}
    \caption{Structure of the two-step renderer pipeline. We use the result of Wav2Lip for reference, U-Net models for reference and denoising, and 3D Morphable Models (3DMM) with pose and blink adjustments.}
    \label{fig:figure3}
    \vspace{-10pt}
\end{figure}

As shown in Figure~\ref{fig:figure3}, the renderer in our system integrates multiple components to generate realistic video frames from source audio and images. Initially, the Wav2Lip~\citep{wav2lip} module aligns lip movements with the input audio, producing a synchronized reference. While Wav2Lip~\citep{wav2lip} serves as the input to the Reference UNet, our key contribution lies in designing a pipeline that generates precise and accurate optical flow, moving beyond simply replicating Wav2Lip~\citep {wav2lip} output. Specifically, we introduce a Reference UNet with self-attention to enhance feature extraction and focus on learning optical flow that guides realistic facial feature movements. Simultaneously, the 3DMM aids in capturing facial dynamics such as pose and blink, facilitated by an existing Mapping Net. The outputs from these stages are fed into a Denoising UNet, which predicts the optical flow necessary for the warping mechanism to generate new frames. Finally, the frames are enhanced by RefineNet to ensure high-quality and coherent video output. Crucially, the novelty of our system lies in the integration of the flow-based warping mechanism and the refinement of optical flow predictions to better capture natural and synchronized audio-visual dynamics. By leveraging cross-attention mechanisms to effectively integrate audio-visual cues and 3DMM data, our method ensures a seamless and realistic rendering process.

\subsection{Training Loss} 
In addition to the noise prediction loss in Equation~\ref{equ:noise_loss}, we also incorporate the velocity loss $\mathcal{L}{v}$. Initially, we derive the predicted motion $\hat{\mathbf{x}_t}$ from the estimated noise $\hat{\mathbf{\epsilon}_t}$. Then, we have velocity loss computed as the mean square error of the velocity:
\begin{equation}
\mathcal{L}_{v} = \mathbb{E}\left[\left\|(\mathbf{x}_0[1:] - \mathbf{x}_0[:-1]) - (\hat{\mathbf{x}}_0[1:] - \hat{\mathbf{x}}_0[:-1]) \right\|^{2}\right]
\end{equation}

\begin{figure*}[h]
    \centering    \includegraphics[width=1\linewidth]{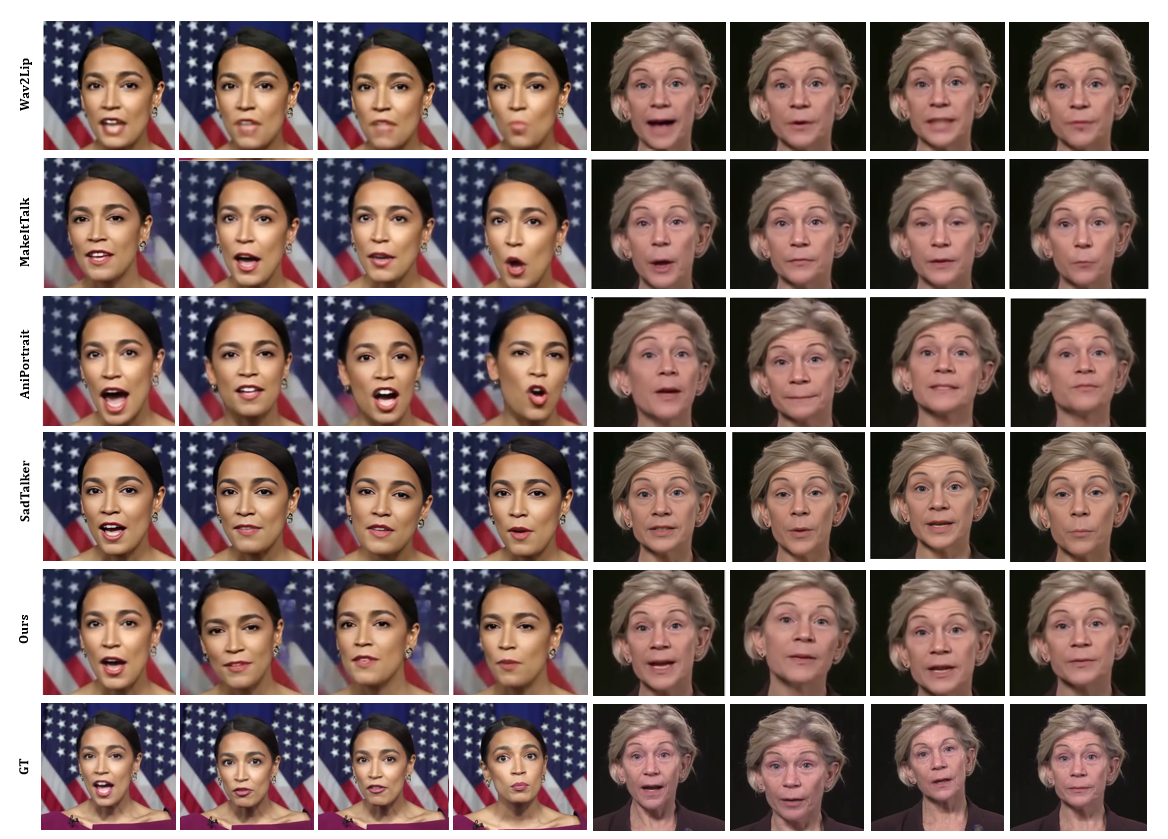}
    \caption{Comparison of lip-syncing models: Visual results for Wav2Lip~\citep{wav2lip}, MakeItTalk~\citep{zhou2020makelttalk}, AniPortrait~\citep{wei2024aniportrait}, SadTalker~\citep{sadtalker}, and our proposed method, compared to the ground truth (GT). Note that only the results for SadTalker use GFPGAN enhancement. More qualitative generation results are shown in the supplementary document.}
    \label{fig:compare}
\end{figure*}

\begin{table*}[h]
	\centering
    \resizebox{\textwidth}{!}{
	\begin{tabular}{l ccc|ccc|ccc|ccc}
\toprule
Dataset              &\multicolumn{6}{c}{HDTF}                        & \multicolumn{6}{c}{VFHQ}\\
\hline
Identity&\multicolumn{3}{c}{Same-Identity}                        &\multicolumn{3}{c}{Cross-Identity}
&\multicolumn{3}{c}{Same-Identity}                        & \multicolumn{3}{c}{Cross-Identity}  
\\ \midrule
& LSE-C$\uparrow$               & LSE-D$\downarrow$             & FID$\downarrow$            & LSE-C$\uparrow$              & LSE-D$\downarrow$              & FID$\downarrow$      & LSE-C$\uparrow$               & LSE-D$\downarrow$             & FID$\downarrow$            & LSE-C$\uparrow$              & LSE-D$\downarrow$              & FID$\downarrow$      \\
\hline
Wav2Lip~\citep{wav2lip}                               &  9.652 & 5.656 & 20.01               & 9.537  & 5.724 & 23.80  &7.982 & 6.964 & 25.10 & 8.073	& 6.958	& 25.56               \\
MakeItTalk~\citep{zhou2020makelttalk}                            & 5.043  & 9.861 & 22.39                & 4.968  & 9.936 & 27.52 &4.074 & 9.542 & 28.75 & 4.019	& 10.574	& 28.47              \\
AniPortrait~\citep{wei2024aniportrait}                           & 4.469  & 10.253 & 31.58                & 4.419  & 10.289 & 31.91 &4.061 & 10.673 & 29.52 & 4.479	& 10.398	& 29.94             \\
SadTalker~\citep{sadtalker}                             & 7.175  & 8.076 & 20.55             & 7.122  & 7.929 & 23.12  &6.108 & 7.949  & 25.72 & 6.179	& 7.082	&   25.07   \\
DreamTalk ~\citep{ma2023dreamtalk}                           & 7.225  & 7.838 & 28.91 &7.098 & 7.932 & 29.17 & 6.053  	&7.648	& 28.75 & 6.371  & 7.359  & 28.82                   \\
Hallo~\citep{xu2024hallohierarchicalaudiodrivenvisual}                           & 7.417  & 7.718 & \textbf{20.54} &\textbf{7.259} & \textbf{7.681} &\textbf{20.68} & 5.867	&7.864	&\textbf{24.53} &\textbf{6.398}&\textbf{7.214} &\textbf{25.93}                    \\
\hline
Ours                                  & \textbf{7.428} & \textbf{7.645} & 22.34                  & 7.251 & 7.809 & 22.83   &\textbf{6.205}  & \textbf{7.437}  &26.11   & 6.279 & 7.548  &  26.93        \\
Ground Truth                          & 7.921 & 7.357 & -            & - & - & -  & 6.458 & 7.162 &  - & - & - & -         \\ 
\bottomrule
\end{tabular}
}
\caption{Comparison with the state-of-the-art method on HDTF and VFHQ dataset. We evaluate Wav2Lip~\citep{wav2lip} in the one-shot settings. Wav2Lip~\citep{wav2lip} achieves the best video quality since it only animates the lip region while other regions are the same as the original frame.}
\label{tab:aud_driven}

\end{table*}

\par We also utilize pre-trained lip-reading models from~\citep{ma2022training} to compute the lip reading loss $\mathcal{L}_{read}$ and landmark loss $\mathcal{L}_{lks}$ from~\citep{sadtalker}. Overall, the final loss of ExpNet is:
\begin{equation}
    \mathcal{L}=\lambda_{t} \mathcal{L}_{t}+\lambda_{read} \mathcal{L}_{read} + \lambda_{lks} \mathcal{L}_{lks} + \lambda_{v} \mathcal{L}_{v}
\end{equation}
Where $\lambda_{t}$, $\lambda_{read}$, $\lambda_{lks}$, $\lambda_{v}$ are set to $10$, $0.2$, $0.1$, and $0.1$.

%% file: Content/Chapter/Experiments.tex
\noindent \textbf{Datasets} Following the settings, we used 500 videos from the HDTF and VFHQ Datasets, whose duration is 8 seconds each. For each video, the initial frame was used as input along with the 8-second audio segment. We crop the original videos following previous image animation methods~\citep{fomm} and resize the video to 256$\times$256. We select ~1k aligned videos and audios of 100 subjects to train our model. 

\noindent \textbf{Implementation Details} In our pipeline, we utilize  Hubert-base-ls960 version~\citep{haque2023facexhubert} as the audio encoder, which is a pre-trained model trained on 960 hours of the LibriSpeech~\cite{7178964} dataset. The audio signals are converted into the form of a spectral frequency latent and used for subsequent training. All the 3DMM parameters are extracted through pre-trained deep 3D face reconstruction method~\cite {deng2019accurate}. We perform all the experiments on 2 A100 GPUs. Our model is learned via continuous 12 frames. We implemented a Transformer model with a configuration that includes 1024 dimensions for hidden states, 2048 dimensions for feedforward networks, alongside 4 heads for multi-head attention, and a single Transformer block. The self/cross-attention in the Transformer utilizes residual connections, meaning that the output of each attention operation is summed with its input. Throughout training sessions, we employed the AdamW optimization algorithm, setting the learning rate to $1e^{-4}$, and the diffusion step is 10k.

\noindent \textbf{Evaluation Metrics}
We demonstrate the superiority of our method on multiple metrics that have been widely used in previous studies. We evaluate the perceptual differences of the mouth shape from Wav2Lip~\citep{wav2lip}, including the distance score~(LSE-D) and confidence score~(LSE-C). We also employ Frechet Inception Distance (FID) ~\citep{fid,fidpaper}.

\subsection{Compare with Other State-of-the-art Methods}

In our primary analysis, we conduct the same-identity experiment. The first frame of the test video is used as the reference image, and the corresponding audio functions as the driving signal, creating a video with aligned expressions but different head movements. Further, in the cross-identity experiment, the driving audio is sourced from another video. This method is frequently employed in comparisons of video-driven face reenactment~\citep{fomm}.

\par We compare our method with several state-of-the-art methods with different approaches: audio to expression generations (Wav2Lip~\citep{wav2lip}), talking head generation  (MakeItTalk~\citep{zhou2020makelttalk}), 3DMM-based talking head generation 
 (SadTalker~\citep{sadtalker}), diffusion-based talking head generation (Aniportrait~\citep{wei2024aniportrait}, Hallo~\citep{xu2024hallohierarchicalaudiodrivenvisual}). As shown in Table~\ref{tab:aud_driven}, Our approach demonstrates the best lip-sync performance across the Same-Identity on both datasets. And it has SOTA-comparable results in Cross-Identity scenarios. To demonstrate the scalability of our model, we evaluated it on a completely out-of-distribution dataset, as shown in Table~\ref{tab:tableHQ}. Our model achieves performance and efficiency comparable to SOTA methods. However, it is important to note that Hallo~\citep{xu2024hallohierarchicalaudiodrivenvisual} outperforms other approaches due to its training on a significantly larger dataset.

\par Additionally, we also visualize some samples to demonstrate the performance of our MoDiT. As shown in Figure~\ref{fig:compare}, we can observe that our method can produce a more accurate and smoother portrait video. These results demonstrate that our method effectively captures spatial-temporal correspondences, enhancing the generation of lip-synced videos and boosting the generation performance with the diffusion transformer model. 

\subsubsection{Testing on Out-of-distribution Dataset}
To demonstrate the scalability of our model, we evaluated it on a completely out-of-distribution dataset, as shown in Table \ref{tab:tableHQ} below. We also added Mouth Landmark Distance~(LMD), and Face Action Unit Error~(AUE) for better demonstrating the spatial and temporal consistency.

\begin{table}[h]
\centering
    \resizebox{0.48\textwidth}{!}{
    \large
    \begin{tabular}{*{10}{c}}
        \toprule
       Model & LSE-C↑ & LSE-D↓ &  FID↓ & LMD↓ &  AUE↓ &  Inference Time \\
        \midrule
        DreamTalk [27] & 7.47 & 7.38 & 25.1 & 2.72 & 2.90 & 4min27s \\
        SadTalker [51]  & 7.54 & 7.33 & 22.5 & \textbf{2.33} & \textbf{2.59} & 2min43s \\
        Hallo [47]      & \textbf{7.69} & \textbf{7.18} & \textbf{20.4} & 2.41 & \underline{2.64} & 5min13s \\
        Ours           & \underline{7.65} & \underline{7.20} & \underline{21.7} & \underline{2.37} & 2.66 & 2min58s \\
        Ground Truth   & 8.39 & 6.57 & - & - & - & - \\
        \bottomrule
    \end{tabular}
    }
    \caption{Evaluation on VoxCeleb-HQ dataset}
    \label{tab:tableHQ}
    \vspace{-10pt}
\end{table}

\subsection{Ablation studies}

\subsubsection{Ablation on pipeline component} 
\begin{figure}[h]
    \centering \includegraphics[width=1\linewidth]{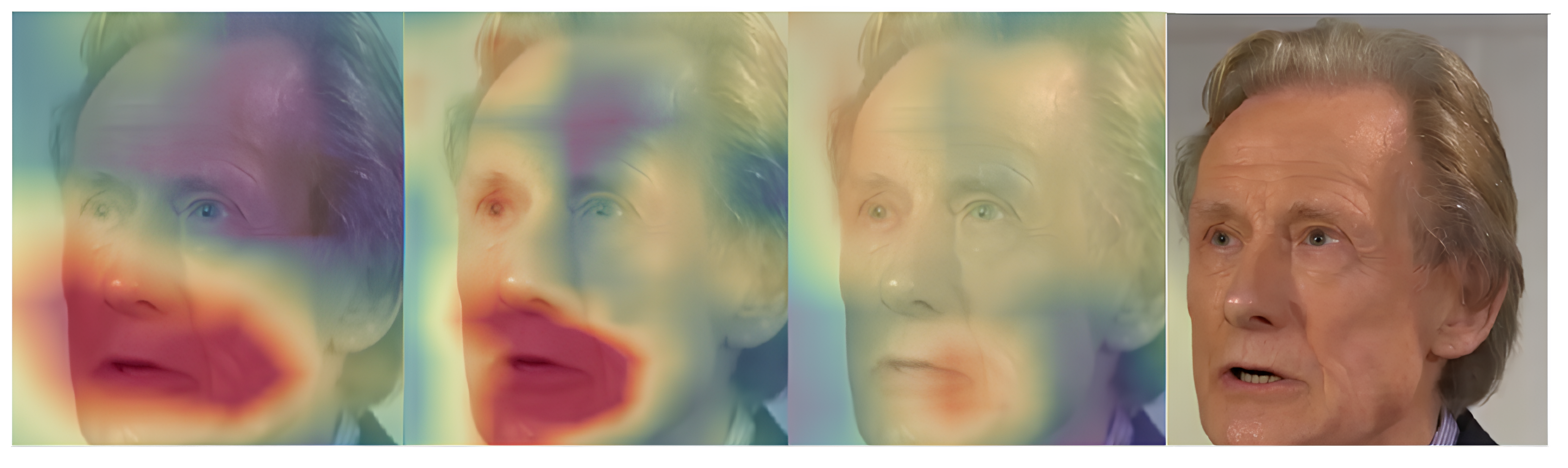}
    \caption{This figure visualizes the attention heatmap shifts over time(initial, middle, and final) with bias injection strategy.}
    \label{fig:blink}
    \vspace{-10pt}
\end{figure}

In our ablation study on the HDTF dataset (Table~\ref{tab:abla}), we evaluated the impact of key components in our model on spatial and temporal coherence. Removing the 3DMM coefficient constraints $\mathbf{\beta}_0$ significantly reduced spatial coherence and affected lip synchronization (low LSE-D, high LSE-C), emphasizing 3DMM's role in maintaining spatial and temporal alignment. Excluding the bias injection strategy impaired spatial alignment and local lip focus, as shown in attention heatmap shifts (Figure~\ref{fig:blink}), highlighting its importance in guiding attention across facial regions during generation. Omitting temporal attention slightly reduced frame-to-frame consistency, indicating its role in ensuring fluid animations, while the addition of $\mathcal{L}{v}$ improved coherence by controlling velocity variations for smoother transitions.

\begin{table}[h]
	\centering
	\footnotesize
	\begin{tabular}{l cc}
        \toprule
& LSE-C$\uparrow$               & LSE-D$\downarrow$\\
\hline
w/o 3DMM source $\mathbf{\beta}_0$                       & 5.365 & 9.431    \\

w/o bias injection                           & 6.994 & 7.942    \\

w/o revised temporal attention                       & 7.265 & 7.801    \\
w/o $\mathcal{L}_{v}$                       
&7.395 & 7.673    \\
\hline
Full                                  & \textbf{7.428} & \textbf{7.645}\\ \bottomrule
\vspace{-10pt}
\end{tabular}
\caption{Ablation studies of pipeline component on HDTF dataset. This table evaluates the impact of excluding the 3DMM source, bias injection, temporal attention, and velocity loss.}
\label{tab:abla}
\end{table}

\subsubsection{Ablation on Renderer} 

\begin{figure}[h]
    \centering \includegraphics[width=1\linewidth]{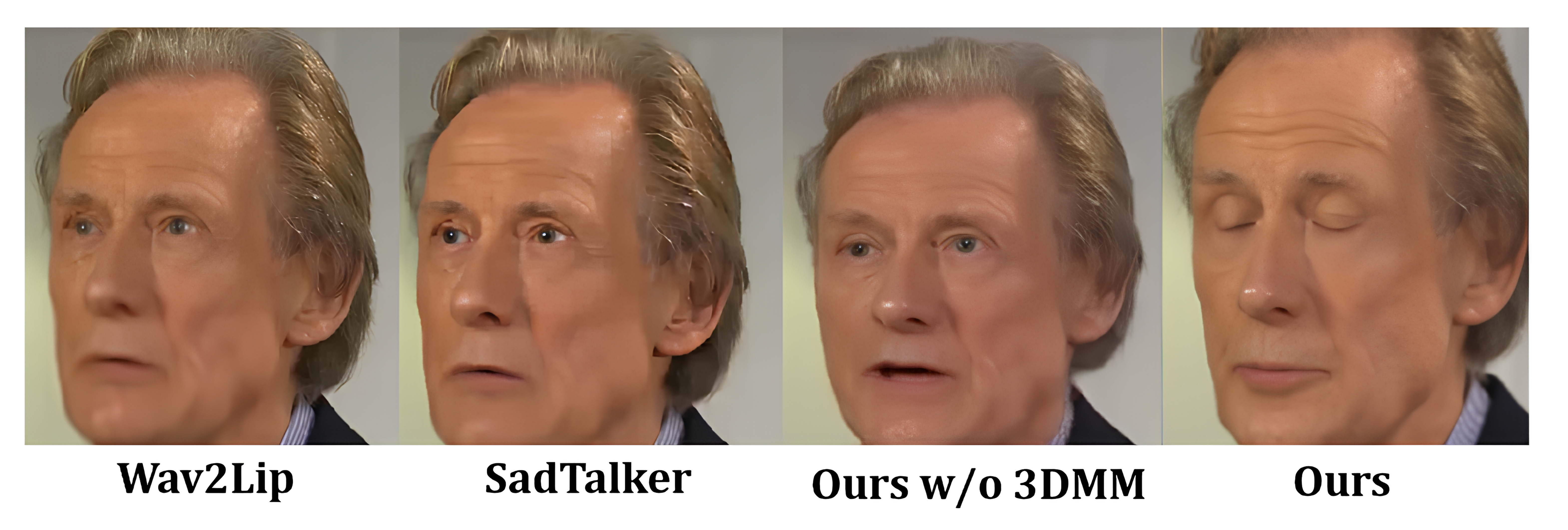}
    \caption{The figure is the ablation study on rendering, comparing with Wav2lip~\citep{wav2lip} and SadTalker~\citep{sadtalker}'s results.}
    \label{fig:blink2}
    \vspace{-10pt}
\end{figure}

Table \ref{tab:abla2} highlights the importance of each component in our model. Removing the 3DMM source or Wav2Lip reference significantly reduced performance, emphasizing their roles in maintaining coherence, precision, and synchronization. Figure~\ref{fig:blink2} also shows comparison results of different renderers. Overall, the full model outperformed the alternatives, demonstrating that the structure and components are both essential for achieving high-quality rendering.

\begin{table}[h]
	\centering
	\footnotesize
    \vspace{10pt}
	\begin{tabular}{l cc}
    \toprule

& LSE-C$\uparrow$               & LSE-D$\downarrow$\\
\hline
w/o 3DMM source $\mathbf{\beta}_0$                       & 1.478 & 13.596    \\

w/o Reference Unet                          & 6.898 & 8.273    \\
\hline
Full                                  & \textbf{7.428} & \textbf{7.645}\\ \bottomrule
\end{tabular}
\caption{Ablation studies of the effect of the renderer. This table evaluates the impact of excluding the 3DMM source and the Reference UNet. }
\label{tab:abla2}
\vspace{-5pt}
\end{table}

In the main table of the experiment, it is evident that Wav2Lip~\cite{wav2lip} achieves strong performance in both LSE-C and LSE-D metrics. This can be attributed to the fact that Wav2Lip focuses exclusively on generating changes in the mouth region. However, the outputs produced by Wav2Lip alone are of relatively low resolution. Similarly, due to its limited scope of generation, Wav2Lip also performs well in the FID metric. Furthermore, when comparing our results with those of Hallo~\cite{xu2024hallohierarchicalaudiodrivenvisual}, it is important to note that the training datasets differ significantly. According to Hallo's paper, their method relies on a large amount of private data, which limits the ability to assess the contribution of their approach in a comparable manner to ours.

\subsubsection{Ablation on eye blink} 

\begin{figure}[h]
    \centering \includegraphics[width=1\linewidth]{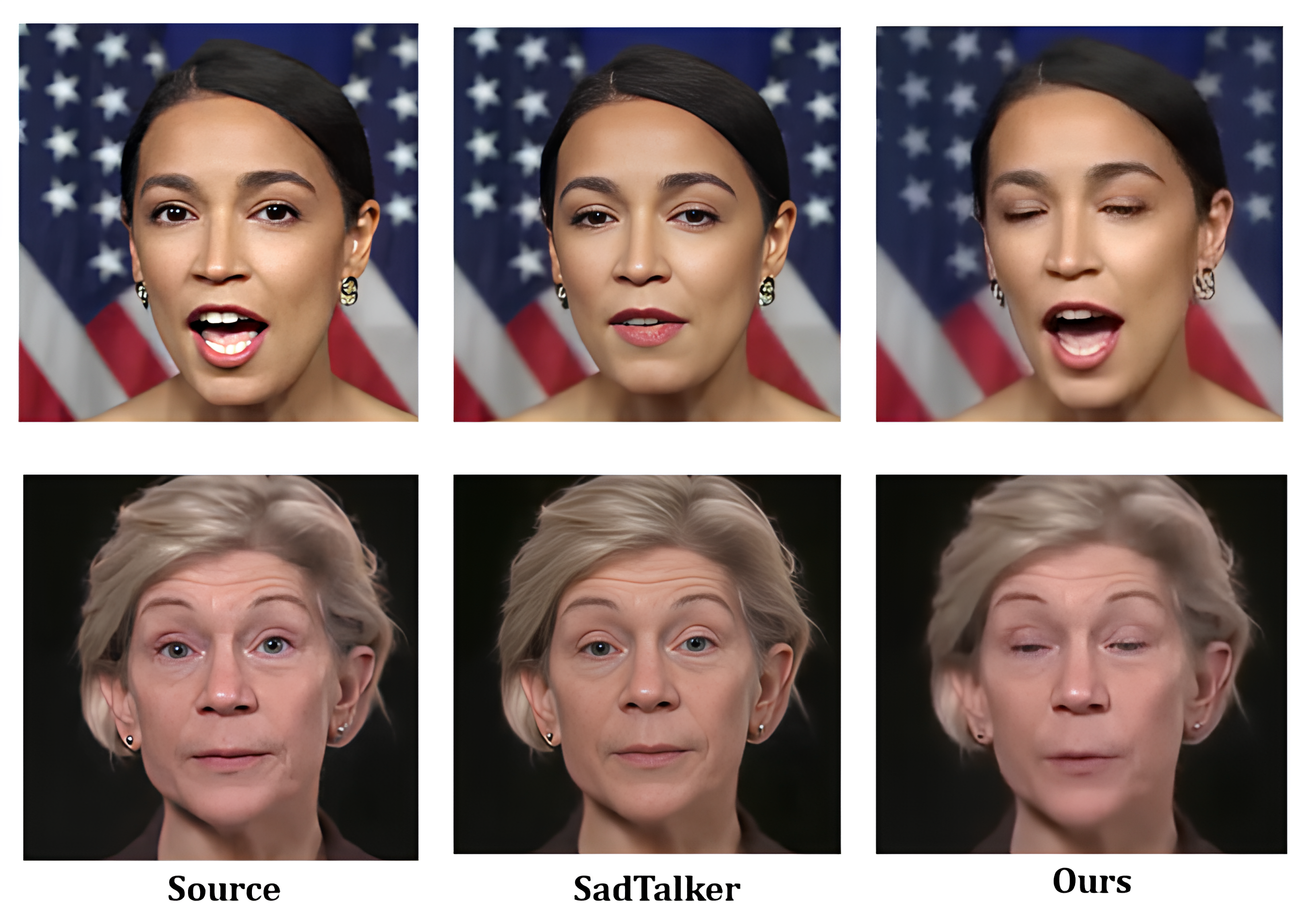}
    \caption{Ablation study on Eye Blink in the source image, maximum blink distance of SadTalker and ours. } 
    \label{fig:blinking}
    \vspace{-10pt}
\end{figure}

In our ablation study, we examined the impact of various components in our eye blink generation model. Removing the Blink Module resulted in a significant drop in the realism of blinks, demonstrating its role in capturing temporal dependencies. We also tested the influence of the learned latent variable by replacing it with a binary signal, which led to less natural blink transitions.

\par In Figure \ref{fig:blinking} below, we present a detailed comparison of the maximum blink distance of eyes in their normal state, utilizing SadTalker~\cite{sadtalker} and our model. Blinking is a rapid process, typically captured in only a few frames, yet our model demonstrates superior performance. Our model achieves a greater maximum blink distance, resulting in a more lifelike and natural-looking blink compared to SadTalker. This improvement enhances the overall realism of the video, providing a more authentic visual experience. In our Supplementary Video (Appendix 6.6), there is also a comparison between different renderers.

%% file: Content/Chapter/Conclusion.tex
In conclusion, MoDiT represents a step forward in the field of audio-driven talking head generation by integrating the 3D Morphable Model with a Diffusion-based Transformer framework. This approach seeks to address challenges such as spatial and temporal inconsistencies, with promising results in lip synchronization and the realism of animations. By leveraging 3DMM for spatial control and employing a hierarchical attention strategy, our framework aims to reduce issues like jitter and identity shifts while encouraging natural expressions. Furthermore, the improved blinking strategy contributes to the expressiveness of the generated videos, enhancing their potential for user engagement. Our results, validated on datasets such as Voxceleb, HDTF, and VFHQ, indicate the robustness of the method and its potential to contribute to advancements in digital human technology. While this work addresses some existing limitations, we view it as a foundation for further exploration and innovation in virtual communication technologies. We hope this research inspires future efforts to refine and expand.